\documentclass[runningheads]{llncs}
\usepackage{amsmath,amssymb}
\usepackage{graphicx}
\usepackage{booktabs}    % 三线表
\usepackage{color}  
\usepackage{multirow}% 可选，用于标注颜色
\usepackage{algorithm}
\usepackage{algorithmic}
\usepackage{comment}
\usepackage{multirow}
\usepackage{amssymb}  % if you use checkmarks elsewhere
\usepackage{caption}
\usepackage{subcaption}
\usepackage{array}
\usepackage{adjustbox}

\usepackage{booktabs}
\usepackage{siunitx}    
\usepackage{hyperref} % 开启超链接跳转（放在所有宏包最后）
\definecolor{myblue}{RGB}{60,120,216}
\hypersetup{
  colorlinks=true,
  linkcolor=myblue,
  citecolor=myblue,
  urlcolor=myblue
}

\begin{document}
\title{
% Unified Structural-Semantic Reciprocal Learning for Unsupervised Visible-Infrared Person Re-Identification
Structural-Semantic Reciprocal Learning for Unsupervised Visible-Infrared Person Re-Identification
}
\author{Moyao Tian\inst{1,}$^{\dag}$ \and
Shijia Liu\inst{1,}$^{\dag}$ \and
Yan Yang\inst{3,4} \and 
Xin Yuan\inst{1,2,}$^{*}$
% \thanks{Corresponding author: Xin Yuan (yuanxincherry@gmail.com)} 
\and
Minshi Chen\inst{1} \and \\
Wei Wang\inst{1,2} \and
Xiao Wang\inst{1,2}}
% \footnotetext[0]{$\dag$ Equal contribution.}
% \renewcommand{\thefootnote}{} % 清空脚注编号，删掉前面的0
% \renewcommand{\thefootnote}{123}
% \author{Moyao Tian\inst{1}\orcidID{0000-1111-2222-3333} \and
% \author{Paper ID: 951}
\institute{School of Computer Science and Technology, Wuhan University of Science and Technology, Wuhan 430065, China \and
Hubei Province Key Laboratory of Intelligent Information Processing and Real-Time Industrial System, Wuhan University of Science and Technology, Wuhan 430065, China \and
State Key Laboratory of Robotics and Intelligent Systems, Shenyang Institute of Automation, Chinese Academy of Sciences, Shenyang 110016, China \and
China University of Chinese Academy of Sciences, Beijing 100049, China}

% ========== 核心去0+左对齐代码 ==========
% 先取消脚注编号，消除数字0
\renewcommand{\thefootnote}{}
% 插入无编号、纯†注释，天然左对齐
\footnotetext{$^{\dag}$ Equal contribution}
\footnotetext{$^{*}$ Corresponding author: Xin Yuan (yuanxincherry@gmail.com)}
% ======================================

\maketitle         % typeset the header of the contribution        
\begin{abstract}
Unsupervised visible-infrared person re-identification (USVI-ReID) is challenging due to the large modality gap and the lack of cross-modal identity annotations. Progressive association paradigms have been proposed to gradually bridge the gap, but they suffer from two critical bottlenecks: reliance on ambiguous global representations and unchecked propagation of pseudo-label noise in an open-loop manner. To address these issues, we propose Structural-Semantic Reciprocal Learning (SSRL), a framework that transforms open-loop association into a self-correcting closed-loop system. Structurally, we introduce Fine-grained Structural Decoupling (FSD) to extract discriminative body-part primitives as reliable spatial anchors, complementing ambiguous holistic silhouettes with spatially consistent structural details. Semantically, we design a Closed-loop Semantic Calibration (CSC) mechanism that reconstructs shared semantic prototypes at each epoch and feeds them back into the training loop, effectively filtering pseudo-label noise before the next clustering cycle. Through the reciprocal interaction between structural and semantic learning, SSRL achieves robust cross-modal representation. Extensive experiments demonstrate the competitive performance of SSRL against state-of-the-art USVI-ReID methods on both SYSU-MM01 and RegDB, notably surpassing several supervised counterparts on RegDB.
\end{abstract}
\begin{figure}[h]
\includegraphics[width=\textwidth]{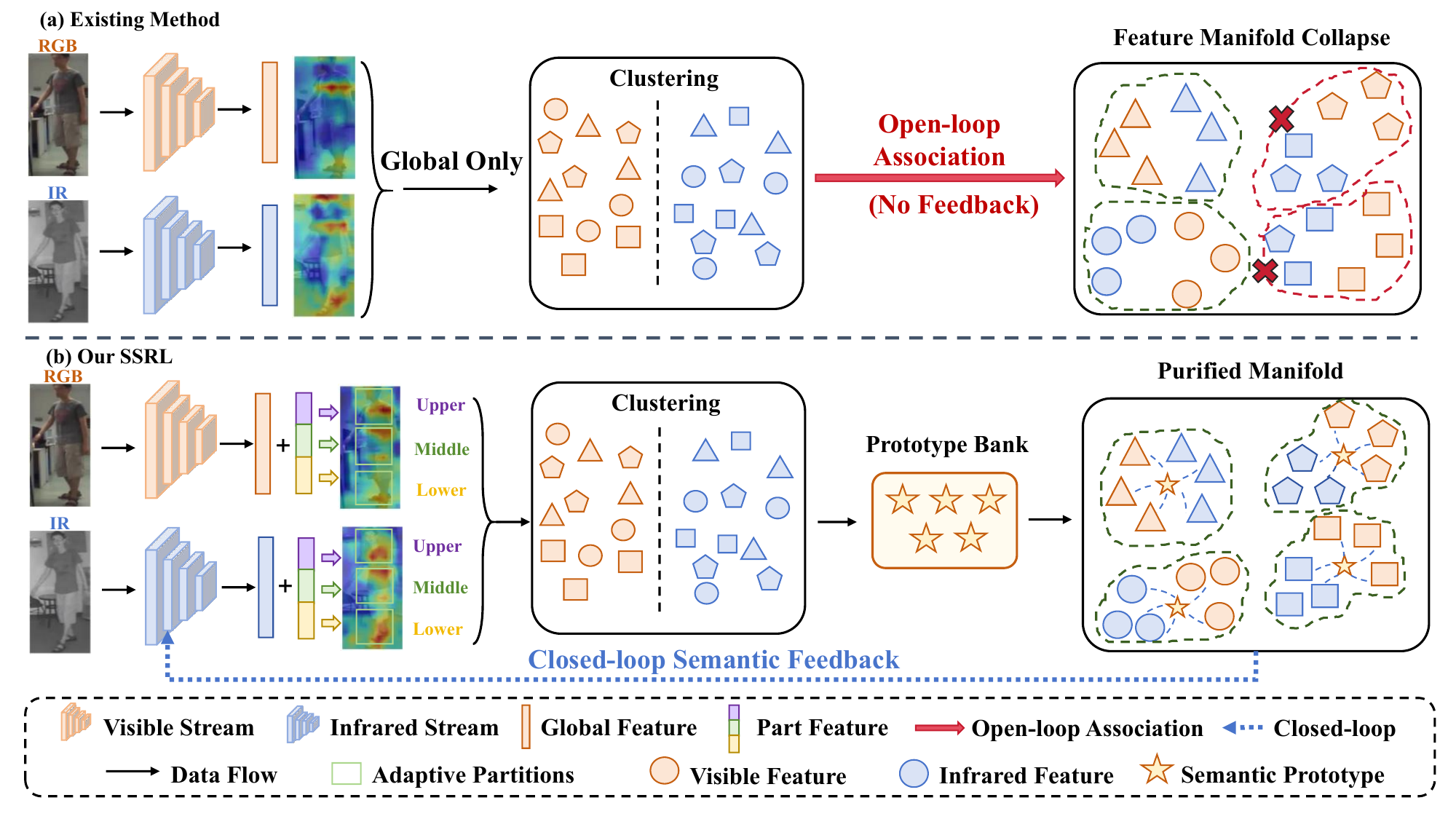}
\caption{Comparison of paradigms. (a) Open-loop methods scatter features; (b) SSRL yields consistent, purified cross-modal clusters.} \label{fig1}
\end{figure}
\section{Introduction}
% Person Re-Identification (Re-ID) retrieves individuals across non-overlapping camera views \cite{1,2}. To address performance degradation in low-light conditions, Visible-Infrared Re-ID (VI-ReID) \cite{5,8} leverages infrared sensors for 24-hour monitoring. Since supervised VI-ReID requires labor-intensive cross-modal annotations, Unsupervised VI-ReID (USL-VI-ReID) \cite{12,14} offers a scalable approach for learning modality-invariant representations without manual labels.
% revised by xinyuan
The goal of visible-infrared person re-identification (VI-ReID) \cite{5,8,34,39} is to match the same pedestrian across visible and infrared images, which is crucial for 24-hour intelligent surveillance and public security. Significant progress has been made in supervised VI-ReID \cite{1,34,35,36,37}. However, these methods rely heavily on manually annotated cross-modal identity labels, which are labor-intensive and expensive to obtain, limiting their applicability in real-world scenarios. To address this limitation, we investigate an unsupervised solution for VI-ReID.

% To bridge the substantial visible-infrared modality gap, recent research has shifted from conventional one-step clustering paradigms \cite{15,16} (e.g., ClusterContrast \cite{16}, H2H \cite{15}) to progressive association frameworks \cite{18,19,20} (e.g., GUR \cite{19}, PCAL \cite{21}). These progressive methods employ curriculum learning to transition from reliable intra-modal discovery to challenging inter-modal identity merging. Concurrently, various association mechanisms, including graph matching \cite{18,22,23}, optimal transport \cite{24,25}, and memory-based aggregation \cite{27,28,31} have been developed to enhance cross-modal identity consistency.
% Nevertheless, current frameworks suffer from two bottlenecks. First, they lack architectural fine-grained decoupling; methods like PPLR \cite{32} and PBE \cite{33} treat local cues merely as auxiliary loss items, failing to provide independent structural support against ambiguous holistic silhouettes. Second, the association follows an open-loop pipeline where pseudo-label noise propagates unchecked \cite{19,20,28}. Lacking feedback, accumulated biases inevitably cause feature manifold collapse and severe identity mis-linking.
% revised by xinyuan
For unsupervised single-modality ReID, existing works \cite{15,16,32} utilize clustering to generate pseudo-labels and learn discriminative representations within a homogeneous space. However, in the visible-infrared heterogeneous space, the large modality gap makes it difficult to maintain feature and semantic consistency across modalities. Specifically, cross-modality differences often exceed intra-modality inter-class variations, causing conventional clustering methods to fail in establishing reliable cross-modal links. Nonetheless, cross-modal correspondences play a critical role in bridging the modality gap \cite{19,21,24}. Without reliable correspondences, the model can hardly learn modality-invariant features.

% revised by xinyuan
Recent efforts \cite{18,19,20,28,33} have attempted to mine cross-modal correspondences in an unsupervised manner. Most of these methods adopt progressive association paradigms, which start from reliable intra-modal clustering and gradually merge inter-modal identities. However, they suffer from two critical bottlenecks. First, they predominantly rely on monolithic global representations, neglecting fine-grained local structural details. As shown in Fig. \ref{fig1}(a), global features are vulnerable to modality-induced silhouette distortions and background clutter. Second, their association pipeline is open-loop: pseudo-label noise generated during clustering propagates unchecked into subsequent training epochs, causing feature manifold collapse and severe identity mis-linking. 

% revised by xinyuan
To address these issues, we propose a Structural-Semantic Reciprocal Learning (SSRL) framework that transforms conventional open-loop association into a self-correcting closed-loop system (Fig. \ref{fig1}(b)). Our framework is characterized by two synergistic designs. 
%Structurally, we introduce a Fine-grained Structural Decoupling (FSD) module. Unlike existing part-based methods that treat local cues as auxiliary loss items \cite{28,32}, FSD deploys three independent bottleneck heads at the backbone level to map distinct body-part primitives (upper, middle, lower). These part-level primitives serve as stable spatial anchors against cross-modal distortions, ensuring that cross-modal correspondences are grounded in spatially consistent structural evidence rather than ambiguous holistic silhouettes. 
Structurally, we introduce Fine-grained Structural Decoupling (FSD) to extract body-part primitives as reliable spatial anchors. Unlike existing part-based methods that simply partition features to enrich representation diversity \cite{28,32}, FSD uniquely projects local regions into independent, decoupled sub-spaces prior to clustering. In unsupervised cross-modal scenarios, this architectural constraint prevents localized modality style noise from corrupting other identity cues, regularizing subsequent pseudo-label learning to maintain strict cross-modal structural consistency.
Semantically, we design a Closed-loop Semantic Calibration (CSC) mechanism. Unlike prior open-loop paradigms where label noise accumulates catastrophically, CSC establishes a feedback loop: at the end of each epoch, we reconstruct shared semantic prototypes from the current feature distribution and feed them back into the training loop. This prototype-based calibration acts as a semantic filter that purifies the feature manifold and suppresses pseudo-label noise before the next clustering cycle. The reciprocal interaction between structural anchoring and semantic cleaning progressively refines both the feature representation and the supervisory signals. The main contributions of this paper can be summarized as follows:

% To address these challenges, we propose Unified Structural-Semantic Reciprocal Learning (USSRL), transforming conventional open-loop association into a self-correcting closed-loop system (Fig. \ref{fig1}). Unlike existing methods \cite{28,32} plagued by global representation ambiguity and noise propagation, USSRL leverages the synergy between localized structural evidence and global semantic consistency. Structurally, a Fine-grained Structural Decoupling (FSD) module deploys parallel bottleneck heads at the backbone level to map distinct body-part primitives (upper, middle, lower) independently, reinforcing cross-modal spatial consistency against holistic silhouette distortions. Semantically, a Closed-loop Semantic Calibration (CSC) mechanism disrupts error propagation by establishing a feedback loop that aligns features with shared semantic prototypes at each epoch's conclusion. Acting as a semantic filter, CSC purifies the feature manifold and recalibrates deviating samples, suppressing pseudo-label noise before subsequent clustering. In summary, our main contributions as follows:

\begin{itemize}

    \item 
    % We propose the USSRL framework, which achieves a paradigm shift from ``open-loop association'' to a ``closed-loop system'' through reciprocal structural semantic learning.
    We propose a SSRL framework that transforms conventional open-loop progressive association into a closed-loop self-correcting system, effectively suppressing pseudo-label noise accumulation through reciprocal interaction between structural and semantic learning.

    \item 
    % We introduce the FSD module, which extracts independent structural primitives via parallel bottleneck heads, providing more stable physical anchors for cross-modal association than traditional auxiliary enhancement methods.
    % We introduce the FSD module, which extracts independent structural primitives via parallel bottleneck heads, providing more stable and discriminative physical anchors for cross-modal association than traditional global or auxiliary part-based methods.
    % We introduce the FSD module, which formulates body-part representations as independent structural primitives and integrates them into progressive cross-modal pseudo-label learning, providing more stable and discriminative structural evidence for cross-modal identity association than traditional global or auxiliary part-based methods.
    We introduce the FSD module, which decouples body-part representations into structural primitives serving as reliable anchors for cross-modal pseudo-label learning under severe modality discrepancy.
    \item 
    % We design the CSC mechanism to establish a self-calibration loop using shared semantic prototypes, effectively suppressing noise accumulation in progressive learning.
    We design the CSC mechanism to establish a self-calibration loop using shared semantic prototypes, effectively suppressing noise accumulation in progressive learning without introducing extra inference cost.
    \item 
    % Extensive experiments on the SYSU-MM01 and RegDB datasets demonstrate that USSRL delivers favorable performance compared to existing methods, proving that the integration of architectural structural precision and closed-loop semantic feedback is essential for robust unsupervised cross-modal retrieval.
    Extensive experiments demonstrate the remarkable advantages of SSRL over state-of-the-art unsupervised benchmarks, notably surpassing several supervised counterparts on RegDB, proving that the integration of architectural structural precision and closed-loop semantic feedback is essential for robust unsupervised cross-modal retrieval.
\end{itemize}

\section{Related Work}
\subsection{Supervised Visible-Infrared Person Re-Identification}

Supervised visible-infrared person re-identification (SVI-ReID) aims to match pedestrian identities across visible (RGB) and infrared (IR) modalities under full identity supervision. The primary challenge lies in severe modality discrepancy caused by heterogeneous imaging characteristics. Early SVI-ReID methods mainly learned modality-shared representations through global feature alignment. AGW \cite{1} introduced one of the earliest VI-ReID benchmarks with a two-stream framework, while CA \cite{34} and DEEN \cite{35} further improved global representations via channel enhancement and diverse embedding expansion.

However, holistic global representations alone are insufficient under severe modality-induced feature drift, motivating structural-aware and fine-grained representation learning. PartMix \cite{36} introduced part discovery regularization for local alignment, while PM-WGCN \cite{37} modeled structural relationships through polymorphic masks and wavelet graph convolution, highlighting the importance of local semantic consistency. More recently, transformer-based methods \cite{38,39} demonstrated strong capability in modeling global contextual dependencies and cross-modal interactions. Despite these advances, supervised VI-ReID methods still heavily rely on large-scale identity annotations, motivating growing interest in unsupervised cross-modality person re-identification.
\begin{comment}
\subsection{Unsupervised Learning for Person Re-Identification}

Unsupervised person re-identification (USL-ReID) learns discriminative identity representations without manual labels, typically via clustering-based contrastive learning. ICE \cite{40} introduced inter-instance contrastive encoding for unsupervised feature discrimination, while Cluster Contrast \cite{16} advanced cluster-level memory optimization for representation compactness. To mitigate noisy pseudo labels, subsequent methods incorporated refined optimization strategies, such as fine-grained body-region consistency \cite{32}, camera-conditioned constraints \cite{41}, and progressive feature-label purification \cite{27}. Despite these single-modality advances, directly extending them to visible-infrared settings remains problematic due to severe cross-modality discrepancies, semantic inconsistency, and amplified pseudo-label noise. 
\end{comment}
\subsection{Unsupervised Visible-Infrared Person Re-Identification} 
Recent unsupervised VI-ReID methods mainly address severe modality discrepancy through cross-modal association learning. Early one-step clustering methods, such as H2H \cite{15}, directly performed cross-modal matching and pseudo-label learning, but suffered from unstable associations due to the large RGB-IR gap. To improve stability, progressive frameworks were introduced, including graph-based optimization \cite{18}, unified embedding learning \cite{19}, curriculum-style progressive association \cite{21}, and transport-based distribution alignment \cite{24}, enabling more reliable clustering and cross-modal correspondence.

Beyond global association learning, recent studies explored fine-grained modeling and pseudo-label purification. PBE \cite{33} enhanced cross-modal interaction through part-based feature learning, while APPD \cite{28} introduced adaptive pseudo-label purification to mitigate clustering noise. However, existing methods still predominantly rely on holistic representations and open-loop optimization, where local structural cues are insufficiently exploited, and pseudo-label errors may progressively accumulate during iterative clustering. To overcome these limitations, we propose a Structural-Semantic Reciprocal Learning (SSRL) framework, which jointly leverages fine-grained structural decoupling and closed-loop semantic calibration to achieve more robust cross-modal representation learning.

\section{Method}
\subsection{Overview} 
\begin{figure}[]
\includegraphics[width=\textwidth]{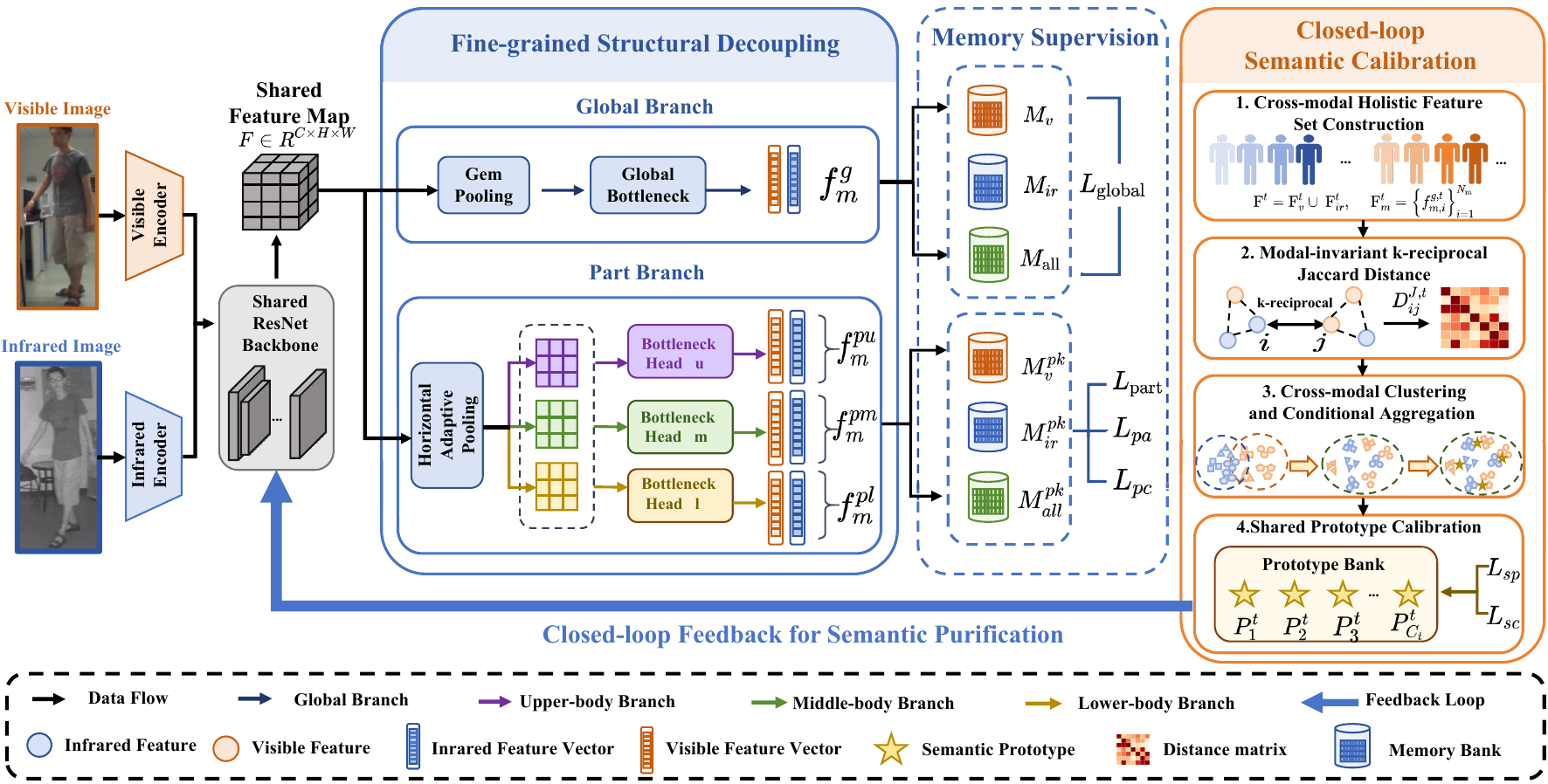}
\caption{Framework of Structural-Semantic Reciprocal Learning (SSRL). SSRL integrates Fine-grained Structural Decoupling (FSD) and Closed-loop Semantic Calibration (CSC) reciprocally. FSD leverages modality-aware memories for spatial feature decoupling, while CSC generates semantic prototypes via cross-modal clustering to progressively refine representation learning.} \label{fig2}
\end{figure}

Given unlabeled visible $\mathbf{X}_v = \{ x_i^v \}_{i=1}^{N_v}$ and infrared $\mathbf{X}_{ir} = \{ x_j^{ir} \}_{j=1}^{N_{ir}}$ images, unsupervised visible-infrared person re-identification learns modality-invariant representations without identity labels. To mitigate the modality gap and pseudo-label noise, we propose Structural-Semantic Reciprocal Learning (SSRL), as shown in Fig.~\ref{fig2}, which synergizes architecture-level Fine-grained Structural Decoupling (FSD) and algorithm-level Closed-loop Semantic Calibration (CSC). FSD extracts global features and local structural primitives as spatial anchors, while CSC leverages cross-modal semantic prototypes to iteratively rectify pseudo-supervision.Formally, an input image $x_m$ ($m \in \{v, ir\}$) is processed by a shallow encoder $E_m(\cdot)$ and a shared backbone $B(\cdot)$ to extract a shared feature map $A_m = B(E_m(x_m)) \in \mathbb{R}^{C \times H \times W}$. FSD then maps $A_m$ into a global feature $f_m^g \in \mathbb{R}^D$ and $K$ part primitives $\mathcal{F}_m^p = \{ f_m^{pk} \}_{k\in \mathcal{K}}$ via $(f_m^g, \mathcal{F}_m^p) = \Phi_{\text{FSD}}(A_m)$, where $\mathcal{K} = \{u, m, l\}$ indexes the upper, middle, and lower part heads. The framework is optimized via an epoch-wise reciprocal loop:
\begin{equation}
\theta_t \xrightarrow{\Phi_{\text{FSD}}} \mathcal{F}^t \xrightarrow{\text{Clustering}} \mathcal{Y}^t \xrightarrow{\text{Aggregation}} \mathcal{P}^t \xrightarrow{\mathcal{L}_{\text{total}}} \theta_{t+1},
\label{eq:1}
\end{equation}
where $\theta_t$, $\mathcal{F}^t$, $\mathcal{Y}^t$, and $\mathcal{P}^t$ denote the model parameters, extracted cross-modal features, synchronized pseudo-labels, and shared prototypes at epoch $t$, respectively. This reciprocal interaction simultaneously stabilizes structural primitives and purifies semantic supervision.

\subsection{Fine-grained Structural Decoupling} 
\noindent\textbf{Global Identity Branch.}
The global branch extracts holistic identity characteristics from the shared feature map $A_m \in \mathbb{R}^{C \times H \times W}$. To retain finer spatial activations, Generalized-Mean (GeM) pooling is adopted to aggregate $A_m$ into a global descriptor $g_m \in \mathbb{R}^C$:
\begin{equation}
g_m = \left( \frac{1}{HW} \sum_{h=1}^{H} \sum_{w=1}^{W} \left( A_m(h,w) \right)^p + \epsilon \right)^{\frac{1}{p}},
\label{eq:2}
\end{equation}
where $p=3$ and $\epsilon$ ensures numerical stability. Then, $g_m$ is projected via a global bottleneck layer $\mathcal{B}_g(\cdot)$ to generate the holistic embedding $f_m^g = \mathcal{B}_g(g_m)$ ($m \in \{v, ir\}$) for instance-level supervision and semantic calibration.

\noindent\textbf{Horizontal Adaptive Pooling.}
To capture spatially stable local cues against background clutter and thermal noise, FSD horizontally partitions $A_m$ into three coarse spatial components. This is executed via an \texttt{AdaptiveAvgPool2d((3,1))} layer, formally denoted as the Horizontal Adaptive Pooling ($\text{HAP}$) operator:
\begin{equation}
p_{m,k} = \frac{1}{|H_k|W} \sum_{h \in H_k} \sum_{w=1}^{W} A_m(h,w), \quad (k \in \mathcal{K}),
\label{eq:3}
\end{equation}
where $H_k$ represents the vertical coordinate range of the segment $k$, and $\mathcal{K} = \{u, m, l\}$ indexes the upper ($u$), middle ($m$), and lower ($l$) body regions. This structural stripe partitioning yields robust cross-modal geometric anchors, inherently providing local spatial invariance against mild pedestrian pose variations.

\noindent\textbf{Parallel Bottleneck Heads.}
To eliminate cross-region feature entanglement and gradient interference, FSD deploys three independent, non-sharing bottleneck heads to process each horizontal bin individually.This process yields the fine-grained structural primitive $k$, denoted as $f_m^{pk} \in \mathbb{R}^{C}$:
\begin{equation}
f_m^{pk} = \mathcal{N}\left(\mathcal{B}_{pk}(p_{m,k})\right), \quad (k \in \mathcal{K}),
\label{eq:4}
\end{equation}
where $\mathcal{B}_{pk}(\cdot)$ denotes the $k$-th part bottleneck and $\mathcal{N}(\cdot)$ represents $L_2$ normalization. The fully decoupled primitive set is represented as $\mathcal{F}_m^p = \{f_m^{pk}\}_{k \in \mathcal{K}}$. Acting as reliable spatial anchors, these primitives provide consistent local evidence under radical modality changes and spatial misalignments.

\noindent\textbf{Part Memory Supervision.}
To enhance part-level discriminability, we introduce a hierarchical part-specific memory supervision loss, $\mathcal{L}_{part} = \mathcal{L}_{part}^{uni} + \mathcal{L}_{part}^{all}$. Specifically, the modality-specific part memory loss $\mathcal{L}_{part}^{uni}$ enforces intra-modality compactness:
\begin{equation}
\mathcal{L}_{part}^{uni} = \sum_{k \in \mathcal{K}} \alpha_k \left[ \mathcal{L}_{nce}\left(f_{v}^{pk}, y_v; M_{v}^{pk}\right) + \mathcal{L}_{nce}\left(f_{ir}^{pk}, y_{ir}; M_{ir}^{pk}\right) \right],
\label{eq:5}
\end{equation}
where $M_{v}^{pk}$ and $M_{ir}^{pk}$ denote non-parametric memories for the $k$-th primitive, $y_v$ and $y_{ir}$ are instance pseudo-labels, and $\mathcal{L}_{nce}(\cdot)$ is the contrastive objective. The cross-modal term $\mathcal{L}_{part}^{all}$ is symmetrically derived using a unified bank $M_{all}^{pk}$ and cross-modal pseudo-labels. The weight configuration $\alpha = [\alpha_u, \alpha_m, \alpha_l] = [1/6, 3/6, 2/6]$ balances the uneven distribution of identity cues across regions (e.g., the torso $m$ exhibits superior texture stability over regions $u$ and $l$ under radical variations).

\noindent\textbf{Cross-modal Structural Alignment.}
To align localized geometric layouts across spectra, we define identical pseudo-label pairs as $\Omega_p = \{(i, j) \mid y_i^v = y_j^{ir}\}$. The cross-modal part alignment loss $\mathcal{L}_{pa}$ is formulated as:
\begin{equation}
\mathcal{L}_{pa} = \sum_{k \in \mathcal{K}} \alpha_k \left[ 1 - \frac{1}{|\Omega_p|} \sum_{(i,j) \in \Omega_p} \text{sim}\left(f_{v}^{pk, i}, f_{ir}^{pk, j}\right) \right],
\label{eq:6}
\end{equation}
where $\text{sim}(\cdot, \cdot)$ denotes cosine similarity.Concurrently, to preserve body topology and prevent structural distortion, we introduce an adjacent-part consistency regularization term $\mathcal{L}_{pc}$ over $\mathcal{F}_m^p$. Instead of full pairwise modeling, $\mathcal{L}_{pc}$ constrains local relations strictly between neighboring regions:
\begin{equation}
\mathcal{L}_{pc} = \frac{1}{2} \sum_{i \in \mathcal{K}} \sum_{j \in \mathcal{K}} \beta_{ij} \left[ \left(1-\text{sim}(f_v^{pi}, f_v^{pj})\right) + \left(1-\text{sim}(f_{ir}^{pi}, f_{ir}^{pj})\right) \right],
\label{eq:7}
\end{equation}
where $i$ and $j$ denote topologically adjacent body parts, and $\text{sim}(\cdot,\cdot)$ is cosine similarity. The term $\beta_{ij}$ represents a weight derived by averaging and normalizing the corresponding region scale priors, where $\beta_{ij}$ is computed as the ratio of the adjacent pair's joint prior $\tilde{\beta}_{ij} = \frac{\alpha_i + \alpha_j}{2}$ to the sum of joint priors across all adjacent pairs. This constraint ensures smooth structural transitions and suppresses spatial drift of part heads during optimization.

\noindent\subsection{Closed-loop Semantic Calibration}
\textbf{Cross-modal Feature Set Construction.}
At the beginning of each epoch $t$, we freeze the model parameters $\theta_t$ to extract holistic representations. To facilitate unified clustering, modality-specific features are aggregated into a cross-modal set $\mathcal{F}^t = \mathcal{F}_v^t \cup \mathcal{F}_{ir}^t$, where $\mathcal{F}_m^t = \{f_{m, i}^{g, t}\}_{i=1}^{N_m}$ ($m \in \{v, ir\}$) and $f_{m, i}^{g, t} \in \mathbb{R}^D$ denotes the holistic feature of the $i$-th sample. This establishes a co-shared embedding space for graph-based label synthesis.

\noindent\textbf{Modal-invariant $k$-reciprocal Jaccard Distance.}
To suppress cross-spectral false positives, the structural similarity between samples $i$ and $j$ is evaluated via a $k$-reciprocal Jaccard distance. For an instance $i \in \mathcal{F}^t$, its stringent $k$-reciprocal neighbor set is defined as $\mathcal{R}_k^t(i) = \{ j \mid j \in \mathcal{N}_k^t(i) \wedge i \in \mathcal{N}_k^t(j) \}$, where $\mathcal{N}_k^t(i)$ denotes the $k$-nearest neighbors. By measuring neighborhood topology consistency rather than absolute distances, the modal-invariant Jaccard distance $D_{ij}^{J, t}$ is formulated as:
\begin{equation}
D_{ij}^{J, t} = 1 - \frac{\left| \mathcal{R}_k^t(i) \cap \mathcal{R}_k^t(j) \right|}{\left| \mathcal{R}_k^t(i) \cup \mathcal{R}_k^t(j) \right|}.
\label{eq:8}
\end{equation}
% Density-Based Spatial Clustering of Applications with Noise (DBSCAN) is subsequently deployed on $D^{J, t}$ to generate the synchronized pseudo-label set $\mathcal{Y}^t = \{ y_i^t \mid i \in \{1, 2, \dots, N_v + N_{ir}\} \} = \text{DBSCAN}(D^{J, t})$,thereby filtering out modality discrepancies.
Density-Based Spatial Clustering of Applications with Noise (DBSCAN) is subsequently applied to $D^{J,t}$ to generate the synchronized pseudo-label set$\mathcal{Y}^{t}=\{y_i^{t}\mid i\in\{1,2,\ldots,N_v+N_{ir}\}\}=\mathrm{DBSCAN}(D^{J,t})$, thereby providing reliable pseudo labels for subsequent training.

\noindent\textbf{Shared Semantic Prototype Construction.}
To rectify pseudo-supervision noise, CSC constructs a centralized prototype bank $\mathcal{P}^t = \{ P_c^t \}_{c=1}^{C_t}$ to serve as cross-modal semantic anchors. For each pseudo-class $c$, the visible centroid $\mu_{c, v}^t = \frac{1}{|\mathcal{C}_c^v|} \sum_{i \in \mathcal{C}_c^v} f_{v, i}^{g, t}$ and the infrared centroid $\mu_{c, ir}^t = \frac{1}{|\mathcal{C}_c^{ir}|} \sum_{j \in \mathcal{C}_c^{ir}} f_{ir, j}^{g, t}$ are calculated on their respective sample index subsets $\mathcal{C}_c^v$ and $\mathcal{C}_c^{ir}$. To robustly handle asymmetric modality distributions and cluster explosion, the raw prototype $\bar{P}_c^t$ is dynamically aggregated via a conditional topology constraint:
\begin{equation}
\bar{P}_c^t = \begin{cases}
\frac{1}{2}(\mu_{c, v}^t + \mu_{c, ir}^t), & \mathcal{C}_c^v \neq \emptyset \wedge \mathcal{C}_c^{ir} \neq \emptyset, \\
\mu_{c, v}^t, & \mathcal{C}_c^v \neq \emptyset \wedge \mathcal{C}_c^{ir} = \emptyset, \\
\mu_{c, ir}^t, & \mathcal{C}_c^v = \emptyset \wedge \mathcal{C}_c^{ir} \neq \emptyset,
\end{cases}
\label{eq:9}
\end{equation}
followed by $L_2$-normalization such that $P_c^t = \frac{\bar{P}_c^t} {||\bar{P}_c^t||_2}$. Distinct from conventional momentum or EMA updates, the prototype bank is completely reconstructed at the end of each epoch based on the newly generated pseudo-labels $\mathcal{Y}^t$. This epoch-wise full reconstruction eliminates historical pseudo-label bias and ensures strict alignment between the semantic bank and the current feature distribution.

\noindent\textbf{Backward Semantic Calibration.}
To enforce reciprocity, CSC feeds $\mathcal{P}^t$ back into the backpropagation loop to rectify the feature distribution. For a feature $f_{m, i}^{g, t}$, its posterior probability of aligning with prototype $P_c^t$ is:
\begin{equation}
p(c \mid f_{m, i}^{g, t}) = \frac{\exp\left( (f_{m, i}^{g, t})^\top P_c^t / \tau \right)}{\sum_{q=1}^{C_t} \exp\left( (f_{m, i}^{g, t})^\top P_q^t / \tau \right)},
\label{eq:10}
\end{equation}
where $\tau$ is the temperature. To minimize pseudo-supervision noise and tighten classes, the prototype classification loss $\mathcal{L}_{sp}$ is optimized over visible ($\mathcal{B}_{v} $) and infrared ($\mathcal{B}_{ir}$) mini-batches:
\begin{equation}
\mathcal{L}_{sp} = -\sum_{i \in \mathcal{B}_v} \log p(y_i^v \mid f_{v, i}^{g, t}) - \sum_{j \in \mathcal{B}_{ir}} \log p(y_j^{ir} \mid f_{ir, j}^{g, t}).
\label{eq:11}
\end{equation}
Concurrently, to bound spectral variations, CSC imposes a cross-modal semantic consistency loss: 
\begin{equation}
\mathcal{L}_{sc} = 1 - \frac{1}{|\Omega_s|} \sum_{(i,j) \in \Omega_s} \text{sim}\left(f_{v, i}^{g, t}, f_{ir, j}^{g, t}\right),
\label{eq:12}
\end{equation}
where $\Omega_s = \{(i,j) \mid y_i^v = y_j^{ir}\}$ defines intra-batch pairs sharing identical pseudo-labels. The joint objective for CSC is $\mathcal{L}_{csc} = \lambda_{sp} \mathcal{L}_{sp} + \lambda_{sc} \mathcal{L}_{sc}$. This calibration abstracts an epoch-wise reciprocal feedback state machine: $\theta_t \rightarrow \mathcal{F}^t \rightarrow D^{J, t} \rightarrow \mathcal{Y}^t \rightarrow \mathcal{P}^t \rightarrow \mathcal{L}_{csc}^t \rightarrow \theta_{t+1}$, which progressively purifies supervisory targets, suppresses noise, and structurally calibrates the cross-modal feature manifold.

\subsection{Overall Optimization}
\noindent\textbf{Global Memory Supervision.}
To anchor macro-identity semantics before localized matching, SSRL adopts a non-parametric global contrastive learning paradigm. Symmetrically to the part-level mechanism, the overall global memory loss is defined as $\mathcal{L}_{global} = \mathcal{L}_v + \mathcal{L}_{ir} + \mathcal{L}_{all}^v + \mathcal{L}_{all}^{ir}$, where $\mathcal{L}_v$ and $\mathcal{L}_{ir}$ are NCE constraints over independent modality banks $M_v$ and $M_{ir}$, while $\mathcal{L}_{all}^v$ and $\mathcal{L}_{all}^{ir}$ are consensus penalties computed over a shared bank $M_{all}$ to enforce seamless cross-modal feature alignment.

\noindent\textbf{Final Objective.}
By unifying architecture-level structural decoupling and algorithmic closed-loop calibration, the joint objective $\mathcal{L}_{total}^t$ optimized at training epoch $t$ is formulated as:
\begin{equation}
\mathcal{L}_{total}^t = \mathcal{L}_{global} + \mathcal{L}_{FSD} + \mathcal{L}_{csc} + \lambda_{adv} \mathcal{L}_{adv},
\label{eq:13}
\end{equation}
\begin{comment}
where $\mathcal{L}_{FSD} = \lambda_p \mathcal{L}_{part} + \lambda_{pa} \mathcal{L}_{pa} + \lambda_{pc} \mathcal{L}_{pc}$ and $\mathcal{L}_{csc} = \lambda_{sp} \mathcal{L}_{sp} + \lambda_{sc} \mathcal{L}_{sc}$ are the joint losses derived in previous sections, and $\mathcal{L}_{adv}$ is an auxiliary adversarial loss with its balancing coefficient $\lambda_{adv}$. Driven by $\mathcal{L}_{total}^t$, the model parameters are updated via backpropagation: $\theta_{t+1} = \theta_t - \eta \nabla_{\theta} \mathcal{L}_{total}^t$, where $\eta$ denotes the learning rate and $\nabla_{\theta}$ denotes gradient operator .
\end{comment}
where $\mathcal{L}_{FSD} = \lambda_p \mathcal{L}_{part} + \lambda_{pa} \mathcal{L}_{pa} + \lambda_{pc} \mathcal{L}_{pc}$ and $\mathcal{L}_{csc} = \lambda_{sp} \mathcal{L}_{sp} + \lambda_{sc} \mathcal{L}_{sc}$ are the joint losses derived in previous sections. The term $\mathcal{L}_{adv}$ is an auxiliary triple-modal adversarial alignment loss strictly following the baseline framework~\cite{21}. It introduces a three-class modality classifier and coordinates the channel-augmented (CA) distribution as an intermediate target bridge to align the holistic representations of visible, infrared, and CA domains, where $\lambda_{adv}$ is its balancing coefficient (set to 0.15).Driven by $\mathcal{L}_{total}^t$, the model parameters are updated via backpropagation: $\theta_{t+1} = \theta_t - \eta \nabla_{\theta} \mathcal{L}_{total}^t$, where $\eta$ denotes the learning rate and $\nabla_{\theta}$ denotes gradient operator.

\section{Experiments}
\subsection{Experimental Setup}
\textbf{Datasets and Evaluation Metrics.}
\begin{comment}
We evaluate the proposed USSRL framework on two benchmark datasets:

SYSU-MM01\cite{5}: A large-scale benchmark containing 287,628 visible and 15,792 infrared images of 491 identities captured by 4 visible and 2 near-infrared cameras. The training set includes 395 subjects, and the testing set contains 96 non-overlapping subjects. We test under both All-Search and Indoor-Search modes.

RegDB\cite{51}: A compact dual-modality dataset with 4,120 visible and 4,120 infrared images from 412 subjects. Following standard protocols, 206 identities are used for training and the remaining 206 for testing. We validate both Visible-to-Infrared and Infrared-to-Visible modes.

Evaluation Metrics: Standard Cumulative Matching Characteristic (CMC at Rank-1/10), mean Average Precision (mAP), and mean Inverse Negative Penalty (mINP) are adopted for quantitative assessment.   
\end{comment}
We evaluate SSRL on two benchmark datasets, SYSU-MM01 \cite{5} and RegDB \cite{51}. SYSU-MM01 contains 287,628 visible and 15,792 infrared images of 491 identities captured by 4 visible and 2 infrared cameras, with 395 identities for training and 96 for testing under All-Search and Indoor-Search settings. RegDB includes 4,120 visible and 4,120 infrared images from 412 identities, where 206 identities are used for training and the remaining 206 for testing under Visible-to-Infrared and Infrared-to-Visible protocols. Standard CMC (Rank-1/10), and mAP are adopted for evaluation.

\noindent\textbf{Implementation Details.}
SSRL is implemented in PyTorch with an ImageNet-pretrained ResNet-50 backbone and two modality-specific shallow encoders. All images are resized to $288 \times 144$, and the feature map is processed by a Horizontal Adaptive Pooling layer (\texttt{AdaptiveAvgPool2d((3,1))}) with three independent bottleneck heads. Training is conducted for 50 epochs with a batch size of 64 using the Adam optimizer ($lr=3.5\times10^{-4}$, weight decay $5\times10^{-4}$). A two-stage optimization strategy is adopted: Stage I learns global and structural representations with $\lambda_p=0.3$, $\lambda_{pa}=0.1$, $\lambda_{pc}=0.05$, and $\lambda_{adv}=0.15$, while Stage II activates CSC with $\lambda_{sp}=\lambda_{sc}=0.1$ and $\tau=0.05$. Pseudo-labels are updated at each epoch via DBSCAN clustering with k-reciprocal Jaccard distance, using $(k_1,k_2)=(30,6)$ for intra-modality and $(40,32)$ for cross-modal clustering. The clustering threshold is set to $\epsilon=0.6$ for SYSU-MM01 and $\epsilon=0.3$ for RegDB cross-modal clustering. During inference, auxiliary branches and CSC are discarded, and retrieval uses only the normalized 2048-dimensional global embedding.
\begin{table}[h]
\centering
\caption{
Performance comparison on SYSU-MM01 and RegDB (\%).
The best and second-best results are in bold and bold-underlined, respectively.
}
\label{tab:sota_joint_sysu_regdb}
\footnotesize
\setlength{\tabcolsep}{3.2pt}
\renewcommand{\arraystretch}{1.08}
\resizebox{\textwidth}{!}{
\begin{tabular}{c|c|c||cc|cc||cc|cc}
\hline
\multicolumn{3}{c||}{\textbf{Settings}}
& \multicolumn{4}{c||}{\textbf{SYSU-MM01}}
& \multicolumn{4}{c}{\textbf{RegDB}} \\
\cline{4-11}
\multicolumn{3}{c||}{}
& \multicolumn{2}{c|}{\textbf{All Search}}
& \multicolumn{2}{c||}{\textbf{Indoor Search}}
& \multicolumn{2}{c|}{\textbf{Visible$\rightarrow$Infrared}}
& \multicolumn{2}{c}{\textbf{Infrared$\rightarrow$Visible}} \\
\hline
\textbf{Type} & \textbf{Method} & \textbf{Venue}
& \textbf{Rank-1} & \textbf{mAP}
& \textbf{Rank-1} & \textbf{mAP}
& \textbf{Rank-1} & \textbf{mAP}
& \textbf{Rank-1} & \textbf{mAP} \\
\hline\hline

\multirow{7}{*}{SVI-ReID}
& CAJ$_{+}$ \cite{34} & TPAMI'23 & 71.48 & 68.15 & 78.36 & 81.98 & 85.69 & 79.70 & 84.88 & 78.55 \\
& DEEN \cite{35}      & CVPR'23  & 74.70 & 71.80 & 80.30 & 83.30 & 91.10 & 85.10 & 89.50 & 83.40 \\
& PartMix \cite{36}   & CVPR'23  & 77.78 & 74.62 & 81.52 & 84.38 & 84.93 & 82.52 & 85.66 & 82.27 \\
& PMWGCN \cite{37}    & TIFS'24  & 66.80 & 64.80 & 72.60 & 76.10 & 90.60 & 84.50 & 87.50 & 82.27 \\
& HOS-Net \cite{44}   & AAAI'24  & 75.60 & 74.20 & 84.20 & 86.70 & 94.70 & 90.40 & 93.30 & 89.20 \\
& CSDN \cite{42}      & TMM'25   & 76.70 & 73.00 & 84.50 & 86.80 & 82.00 & 85.00 & 88.50 & 80.40 \\
& AGPI$^2$ \cite{43}  & TIFS'25  & 72.23 & 70.58 & 83.45 & 84.25 & 89.03 & 83.89 & 87.91 & 83.04 \\
\hline

\multirow{11}{*}{USVI-ReID}
& H2H \cite{15}        & TIP'21     & 30.15 & 29.40 & --    & --    & 23.81 & 18.87 & --    & --    \\
& DOTLA \cite{47}      & MM'23      & 50.36 & 47.36 & 53.47 & 51.61 & 85.63 & 76.71 & 82.91 & 74.97 \\
& MBCCM \cite{48}      & MM'23      & 53.14 & 48.16 & 55.21 & 49.83 & 80.79 & 77.87 & 82.82 & 76.93 \\
& CCLNet \cite{45}     & MM'23      & 54.00 & 50.20 & 56.70 & 65.10 & 69.94 & 65.53 & 70.17 & 66.66 \\
& SCA-RCP \cite{46}    & TKDE'24    & 51.41 & 48.52 & 56.77 & 64.19 & 85.59 & 79.12 & 82.41 & 75.73 \\
& IMSL \cite{49}       & TCSVT'24   & 57.96 & \textbf{\underline{53.93}} & 56.90 & 63.67 & 70.08 & 66.30 & 70.67 & 66.35 \\
& BCGM \cite{23}       & MM'24      & \textbf{\underline{58.90}} & 53.60 & 60.30 & 67.00 & 86.10 & 81.10 & \textbf{\underline{86.50}} & \textbf{\underline{81.80}} \\
& SCLNet \cite{29}     & IVC'25     & 55.24 & 49.13 & 57.40 & 63.30 & 73.24 & 66.10 & 74.10 & 65.71 \\
& NLDC \cite{50}       & ICASSP'25  & 57.09 & 51.02 & 58.24 & 65.05 & 84.03 & 78.34 & 80.05 & 75.32 \\
& PCAL$^{+}$ \cite{21} & TIFS'25    & 57.94 & 52.85 & 60.07 & 66.73 & \textbf{\underline{86.43}} & \textbf{\underline{82.51}} & 86.21 & 81.23 \\
& CARR \cite{52}       & ASOC'26  & 55.37 & 53.89 & \textbf{\underline{62.70}} & \textbf{\underline{68.73}} & 75.19 & 69.83 & 75.83 & 68.68 \\
\cline{2-11}

& \textbf{SSRL (Ours)} & This work
& \textbf{59.47} & \textbf{55.35}
& \textbf{63.83} & \textbf{70.07}
& \textbf{92.47} & \textbf{89.42}
& \textbf{91.06} & \textbf{88.19} \\
\hline
\end{tabular}
}
\end{table}

\subsection{Comparison with State-of-the-Art Methods}
We evaluate SSRL against state-of-the-art methods under two paradigms: fully supervised (SVI-ReID) and unsupervised visible-infrared ReID (USVI-ReID). Quantitative results are summarized in Table \ref{tab:sota_joint_sysu_regdb} (Here, the superscript $+$ denotes that $\text{PCAL}^{+}$~\cite{21} is re-implemented under its original optimal batch size of 192).
As reported, SSRL delivers competitive results within the USVI-ReID category on both benchmarks. On SYSU-MM01, it achieves $59.47\%$ Rank-1 / $55.35\%$ mAP under All Search, outperforming the advanced competitor $\text{PCAL}^{+}$ \cite{21} by $+1.53\%$ and $+2.50\%$, respectively, while consistently dominating recent architectures like BCGM \cite{23}, SCLNet \cite{29}, NLDC \cite{50} and the recent 2026 baseline CARR \cite{52} under Indoor Search ($63.83\%$ Rank-1 / $70.07\%$ mAP).

% More remarkably, on RegDB, SSRL logs $92.47\%$ Rank-1 under Visible-to-Infrared and $91.06\%$ Rank-1 under Infrared-to-Visible protocols, surpassing advanced USVI-ReID methods by substantial margins. Crucially, SSRL achieves a cross-paradigm breakthrough by outperforming competitive fully supervised models, including CVPR'23's DEEN \cite{35} (by $+1.37\%$ Rank-1), PMWGCN \cite {37} (by $+1.87\%$ Rank-1), and CSDN \cite {42} (by $+4.42\%$ mAP). This superiority demonstrates that while supervised models often overfit labeled modalities, our unsupervised formulation extracts highly regularized, modal-invariant structural primitives with exceptional generalization capability.

More remarkably, on RegDB, USSRL achieves 92.47\% Rank-1 under the Visible$\rightarrow$Infrared protocol and 91.06\% Rank-1 under the Infrared$\rightarrow$Visible protocol, outperforming existing USVI-ReID methods by substantial margins. Under the same reported settings, USSRL also shows competitive performance compared with representative supervised VI-ReID methods. These results demonstrate the effectiveness of USSRL in learning robust modality-invariant representations without manual identity annotations.

\subsection{Ablation Studies}
\noindent\textbf{Effectiveness of Main Collaborative Modules.}
As shown in Table \ref{tab:ablation_joint_no_index}, we evaluate how FSD and CSC contribute. Adding FSD alone improves Rank-1 by $+3.07\%$ on SYSU-MM01 (All Search) and $+1.67\%$ on RegDB (Visible$\rightarrow$Infrared), proving that part-level partitioning reduces cross-modal gaps. Conversely, using CSC alone boosts RegDB Rank-1 to $91.71\%$, but causes minor drops on SYSU-MM01. This is because global calibration requires localized structural constraints to avoid wrong matching under heavy noise. Finally, the full model (Base+FSD+CSC) achieves the best performance across all metrics, confirming that structural anchoring and semantic cleaning mutually reinforce each other.

\begin{table}[]
\centering
\caption{Ablation studies on SYSU-MM01 and RegDB (\%).}
\label{tab:ablation_joint_no_index}
\small

\resizebox{\linewidth}{!}{%
\begin{tabular}{c c c|c c|c c|c c|c c}
\hline
\multicolumn{3}{c|}{\textbf{Components}} &
\multicolumn{4}{c|}{\textbf{SYSU-MM01}} &
\multicolumn{4}{c}{\textbf{RegDB}} \\
\cline{4-11}
\multicolumn{3}{c|}{} &
\multicolumn{2}{c|}{All Search} &
\multicolumn{2}{c|}{Indoor Search} &
\multicolumn{2}{c|}{Visible$\rightarrow$Infrared} &
\multicolumn{2}{c}{Infrared$\rightarrow$Visible} \\
\hline
Base & FSD & CSC & Rank-1 & mAP & Rank-1 & mAP & Rank-1 & mAP & Rank-1 & mAP \\
\hline
\checkmark &  &  & 54.39 & 51.95 & 59.69 & 66.72 & 86.43 & 82.51 & 86.21 & 81.23 \\
\checkmark & \checkmark &  & 57.46 & 54.51 & 62.68 & 69.14 & 88.10 & 85.93 & 85.98 & 84.27 \\
\checkmark &  & \checkmark & 54.61 & 49.39 & 56.75 & 64.38 & 91.71 & 88.42 & 90.61 & 87.18 \\
\checkmark & \checkmark & \checkmark & \textbf{59.47} & \textbf{55.35} & \textbf{63.83} & \textbf{70.07} & \textbf{92.47} & \textbf{89.42} & \textbf{91.06} & \textbf{88.19} \\
\hline
\end{tabular}%
}
\end{table}
\noindent\textbf{Internal Ablations of FSD and CSC.}
As shown in Table \ref{tab:internal_ablation_overall}(a), using only $\mathcal{L}_{part}$ yields 55.55\% Rank-1. Adding $\mathcal{L}_{pa}$ boosts accuracy to 58.75\%, proving localized cross-modal alignment narrows modality gaps. Introducing $\mathcal{L}_{pc}$ achieves the optimum, confirming that preserving body topologies prevents spatial relation drift for stable anchoring. 
Concurrently, Table \ref{tab:internal_ablation_overall}(b) shows that optimizing $\mathcal{L}_{sa}$ or $\mathcal{L}_{sp}$ alone achieves 56.01\% and 57.55\% Rank-1, respectively. Jointly constraining both ($\mathcal{L}_{sp} + \mathcal{L}_{sa}$) achieves peak performance, confirming that feature alignment and prototype cleaning mutually reinforce cross-modal associations.
\begin{table}[]
\centering
\caption{Internal Ablation Experiments on SYSU-MM01(\%).}
\label{tab:internal_ablation_overall}
\small 
\begin{subtable}[t]{0.48\linewidth}
\centering
\setlength{\tabcolsep}{5.5pt}   % 控制横向紧凑
\renewcommand{\arraystretch}{1.10}
\begin{tabular}{c c c|c c}
\hline
\multicolumn{3}{c|}{\textbf{Components}} & \multicolumn{2}{c}{\textbf{All Search}} \\
\cline{1-3}\cline{4-5}
$\mathbf{L_{part}}$ & $\mathbf{L_{pa}}$ & $\mathbf{L_{pc}}$ & Rank-1 & mAP \\
\hline
$\checkmark$ &  &  & 55.55 & 51.76 \\
$\checkmark$ & $\checkmark$ &  & 58.75 & 55.06 \\
$\checkmark$ & $\checkmark$ & $\checkmark$ & \textbf{59.47} & \textbf{55.35} \\
\hline
\end{tabular}
\captionsetup{justification=centering}
\caption{Internal ablation of FSD.}
\label{tab:fsd_internal_ablation}
\end{subtable}
\hfill
\begin{subtable}[t]{0.48\linewidth}
\centering
\setlength{\tabcolsep}{5.5pt}
\renewcommand{\arraystretch}{1.10}
\begin{tabular}{c c|c c}
\hline
\multicolumn{2}{c|}{\textbf{Components}} & \multicolumn{2}{c}{\textbf{All Search}} \\
\cline{1-2}\cline{3-4}
$\mathbf{L_{sp}}$ & $\mathbf{L_{sa}}$ & Rank-1 & mAP \\
\hline
 & $\checkmark$ & 56.01 & 52.22 \\
$\checkmark$ &  & 57.55 & 53.78 \\
$\checkmark$ & $\checkmark$ & \textbf{59.47} & \textbf{55.35} \\
\hline
\end{tabular}
\captionsetup{justification=centering}
\caption{Internal ablation of CSC.}
\label{tab:csc_internal_ablation}
\end{subtable}
\end{table}

\noindent \textbf{Analysis of Part-weight Ratios.} 
To justify the assignment of structural importance, we evaluated different part-weight ratios for the upper, middle, and lower body regions. As reported in Table~\ref{tab:part_weight_ablation}, the configuration of $\alpha = [1/6, 3/6, 2/6]$ consistently achieves the optimal performance, generating 59.47\% of Rank-1 in SYSU-MM01 and 92.47\% of Rank-1 in RegDB. Compared with uniformly distributed weights (1:1:1) or other biased allocations, this design effectively capitalizes on the higher structural stability and semantic consistency of torso-related cues under severe cross-modal appearance discrepancies.
\begin{table}[h]
\centering
\caption{Ablation of part-weight ratios on SYSU-MM01 and RegDB (\%).}
\label{tab:part_weight_ablation}
\small
\resizebox{0.75\linewidth}{!}{%
\begin{tabular}{c|c c c|c c c}
\hline
& \multicolumn{3}{c|}{\textbf{SYSU-MM01}} & \multicolumn{3}{c}{\textbf{RegDB}} \\
\cline{2-7}
$\alpha_u : \alpha_m : \alpha_l$ & Rank-1 & Rank-10 & mAP & Rank-1 & Rank-10 & mAP \\
\hline
1:1:1 &  57.27&  90.22&  53.91& 90.68 & 95.75 & 86.94 \\
1:2:3 & 53.47 & 88.62 & 50.21 & 89.48 & 94.66 & 86.92 \\
\textbf{1:3:2} & \textbf{59.47} & \textbf{91.84} & \textbf{55.35} & \textbf{92.47} & \textbf{97.80} & \textbf{89.42} \\
2:1:3 & 55.88 & 90.60 & 52.73 & 90.31 & 95.61 & 86.37 \\
\hline
\end{tabular}%
}
\end{table}

\subsection{Parameter Sensitivity Analysis}
\begin{figure}[]
\centering
\includegraphics[width=\linewidth]{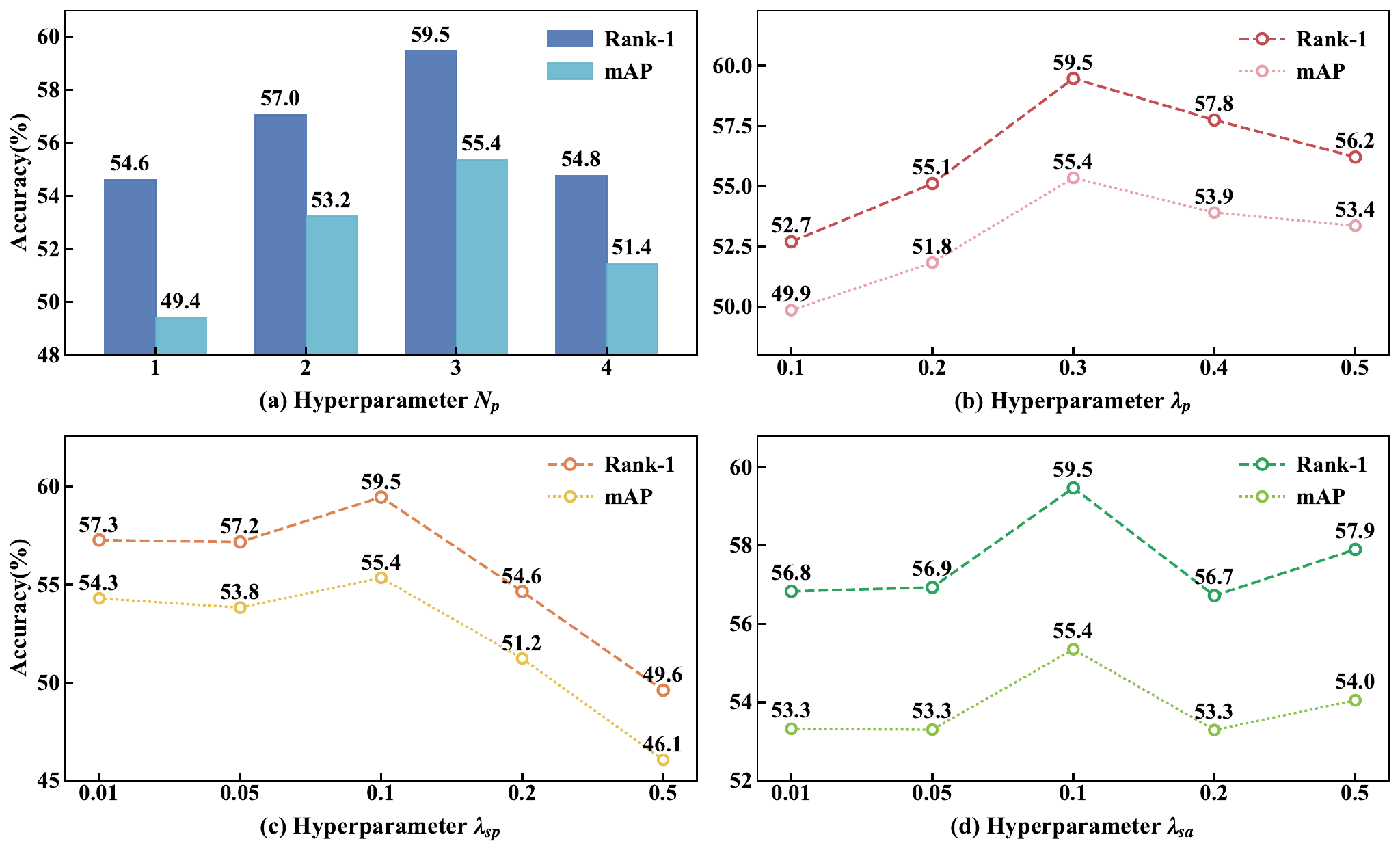}
% -------------- 大图的总 Caption --------------
\caption{Parameter sensitivity analysis on SYSU-MM01(All Search).}
\label{fig:3}
\end{figure}

\noindent\textbf{Impact of Part Partition Cardinality.}
As evaluated in Fig. \ref{fig:3}(a), we analyze performance across different part partitions $N_p \in \{1,2,3,4\}$. Holistic representation ($N_p=1$) yields 54.6\% Rank-1 and 49.4\% mAP. Performance peaks at $N_p=3$ (59.5\% Rank-1 / 55.4\% mAP), confirming that a three-part decomposition optimally aligns with pedestrian anatomical structures. However, over-partitioning ($N_p=4$) disrupts local semantic continuity, degrading results to 54.8\% Rank-1 and 51.4\% mAP.

\noindent\textbf{Sensitivity of Regularization Weights.}
We further evaluate algorithmic robustness by varying the loss hyperparameters $\lambda_p$, $\lambda_{sp}$, and $\lambda_{sa}$ within $[0.01, 0.5]$, as illustrated in Figs. \ref{fig:3}(b)--(d). All three trajectories exhibit a consistent bell-shaped trend, achieving optimal performance symmetrically at $\lambda_p = 0.3$, $\lambda_{sp} = 0.1$, and $\lambda_{sa} = 0.1$. Specifically, smaller coefficients weaken the constraints for local discrimination and noise suppression, whereas excessively large values over-regularize the embedding space, suppressing identity-specific feature representations. The stable performance margins across wide parameter ranges demonstrate the robust calibration capability of the closed-loop learning paradigm.
\subsection{Qualitative Visualizations}
\begin{figure}[h]
\centering
\hspace*{-1.0cm}%
\begin{subfigure}[b]{0.48\linewidth}
  \centering
  \adjustbox{height=4.8cm,valign=M}{\includegraphics{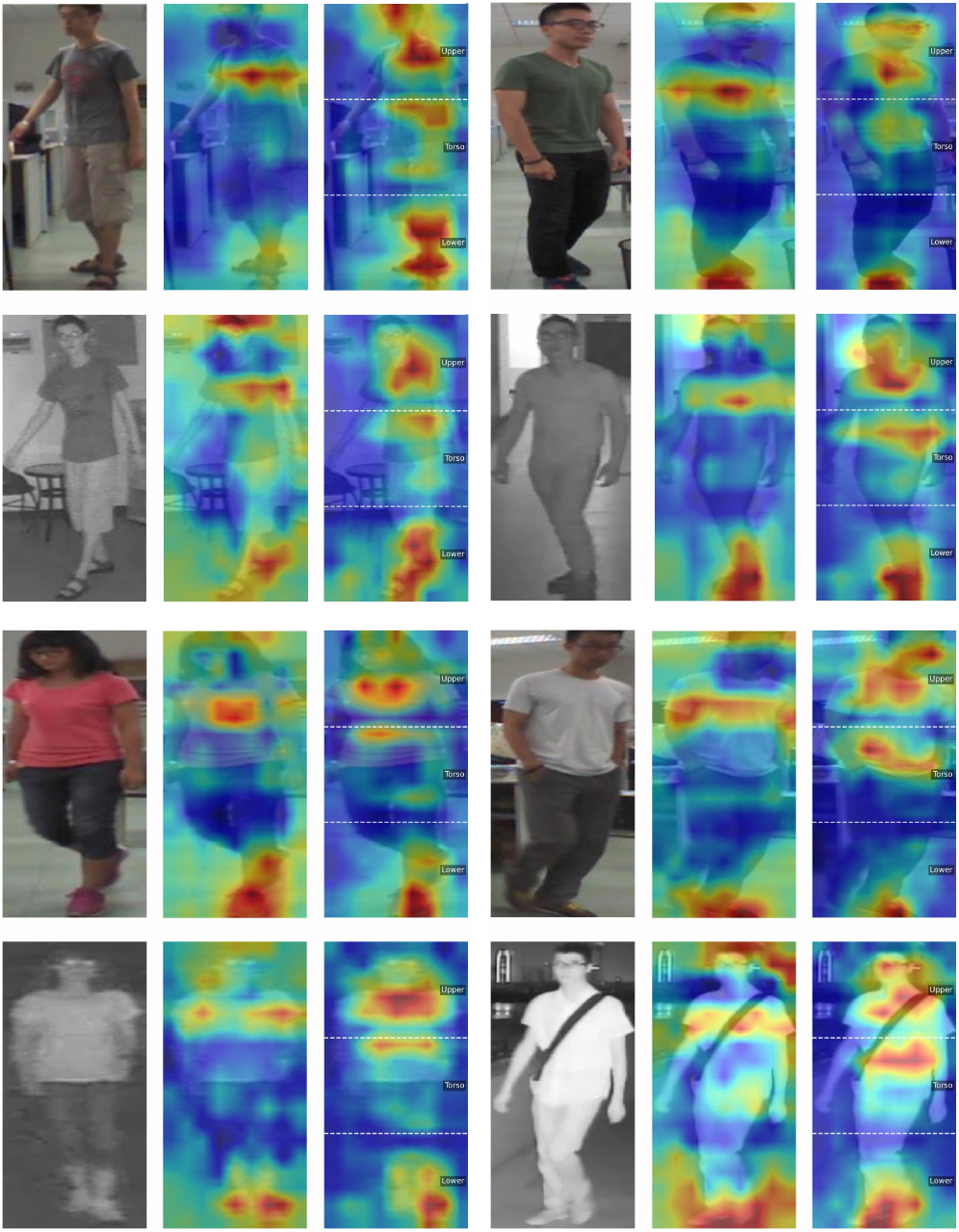}}
  \caption{Grad-CAM on SYSU-MM01.}
  \label{fig:grad_cam_aligned}
\end{subfigure}
\hspace{-0.25cm}
\begin{subfigure}[b]{0.51\linewidth}
  \centering
  \adjustbox{height=4.8cm,valign=M}{\includegraphics{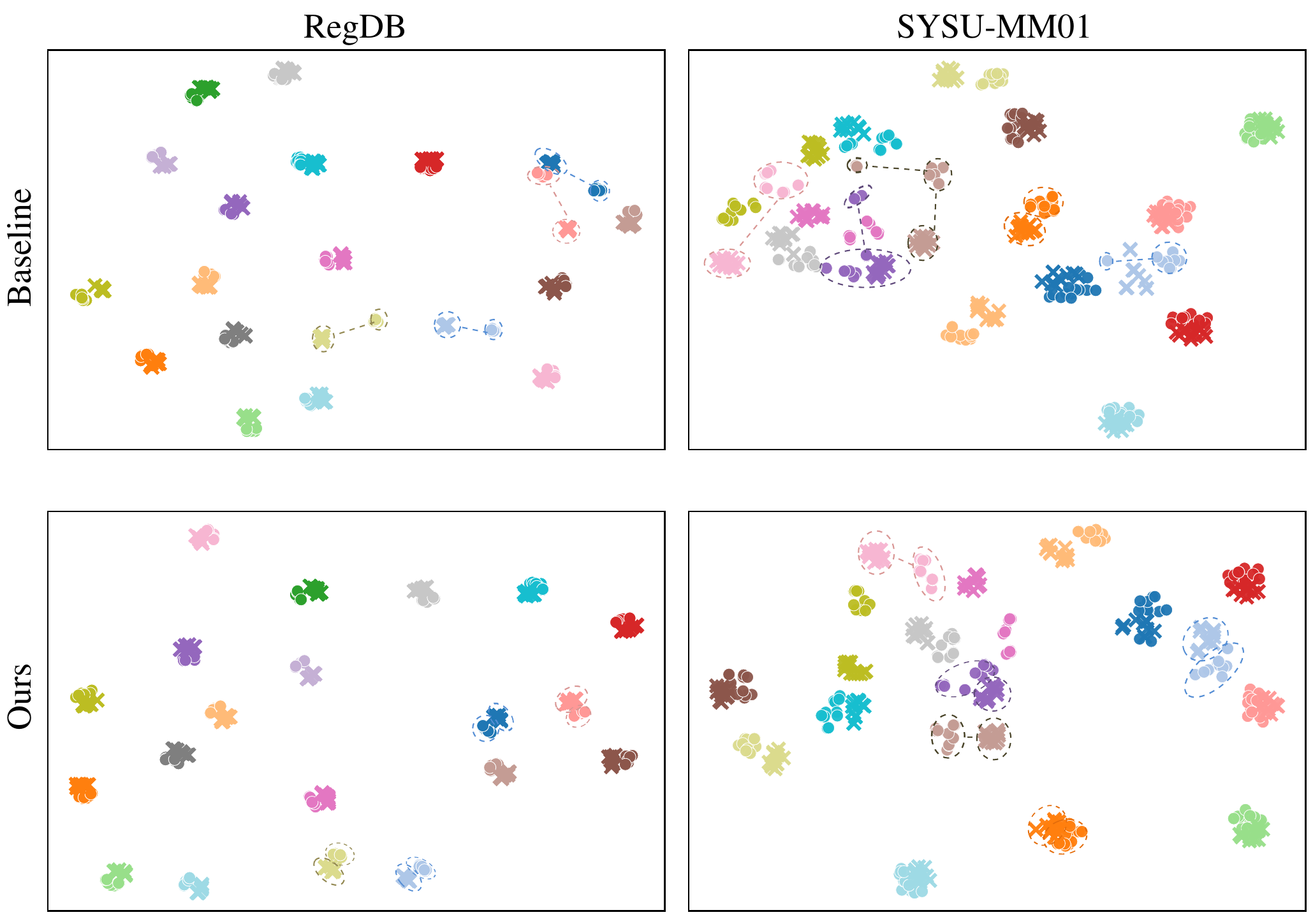}}
  \caption{The t-SNE cluster structures.}
  \label{fig:tsne_aligned}
\end{subfigure}
\caption{Qualitative visualizations of the SSRL.}
\label{fig:qualitative_visualizations}
\end{figure}

\noindent\textbf{Feature Map Activation via Grad-CAM.}
As shown in Fig.~\ref{fig:qualitative_visualizations}(\subref{fig:grad_cam_aligned}), Grad-CAM visualization across four image pairs validates our model's structural robustness. For each triplet (original, baseline, FSD), the baseline exhibits severe modality-sensitive attention drift and background clutter. Conversely, FSD produces symmetric, spectrum-invariant activation maps by consistently focusing on anatomically distinct regions, improving cross-modal consistency and mitigating modality-induced pseudo-label bias.

\begin{comment}
 while the baseline model suffers from chaotic attention drift across modalities and over-focuses on irrelevant background clutter, our FSD module enforces highly symmetric, spectrum-invariant activation maps. Driven by structural primitive constraints, the network precisely anchors its focus onto three anatomically disjoint regions (the upper-body, torso, and lower limbs). This rigid cross-modal localization proves that USSRL successfully isolates identity-associated geometric structures from volatile spectral variations, preventing pseudo-label clustering from shifting toward modality shortcuts.
\end{comment}

% \begin{figure}[]
% \centering
% % [width=\linewidth]
% \includegraphics[width=\linewidth]{fig4.pdf}
% \caption{Grad-CAM activation profiles on SYSU-MM01.}
% \label{fig:gradcam_evidence}
% \end{figure}

\noindent\textbf{Feature Manifold Clustering via t-SNE.}
\begin{comment}
As illustrated in Fig. \ref{fig:tsne_clustering}, we project the 2048-dimensional features of 10 random identities into 2D space to evaluate semantic calibration. In the baseline architecture (a), the embedding space suffers from catastrophic cross-modal fracturing, where visible (circles) and infrared (triangles) samples of the identical identity are split into distant sub-clusters due to the massive modality gap. Symmetrically, in our USSRL model (b), the cross-spectral features sharing the same identity color are tightly and tightly aggregated with clear margins. This sharp contrast demonstrates that the closed-loop semantic calibration successfully eliminates clustering noise and unifies cross-modal distributions into a modal-invariant manifold.
\end{comment}
As illustrated in Fig.~\ref{fig:qualitative_visualizations}(\subref{fig:tsne_aligned}), we project 2048-dimensional features of 20 identities from RegDB and 17 identities from SYSU-MM01 into 2D space for semantic analysis. The baseline model exhibits evident cross-modal fragmentation, where RGB and IR samples of the same identity form separated sub-clusters, whereas SSRL produces compact and well-aligned cross-modal clusters with clearer margins. This demonstrates that the proposed closed-loop semantic calibration effectively suppresses clustering noise and promotes modality-invariant feature learning.

% \begin{figure}[]
% \centering
% \includegraphics[width=0.5\linewidth]{LaTeX2e_Proceedings_Templates/fig5.pdf}
% \caption{Feature manifold clustering via t-SNE on SYSU-MM01. }
% \label{fig:tsne_clustering}
% \end{figure}
\section{Conclusion}
\begin{comment}
In this paper, we present Unified Structural Semantic Reciprocal Learning (USSRL), a novel unsupervised cross-modal person re-identification framework designed to mitigate severe cross-spectral modality gaps and pseudo-label clustering noise. USSRL distinctively orchestrates two collaborative modules: Fine-grained Structural Decoupling (FSD) and Closed-loop Semantic Calibration (CSC). Specifically, FSD partitions the pedestrian body into distinct anatomical segments to capture spectrum-invariant geometric structures, while CSC introduces a robust feedback loop to progressively purify pseudo-labels and align global feature manifolds. Extensive experiments on the SYSU-MM01 and RegDB datasets demonstrate that these modules mutually reinforce each other, allowing USSRL to deliver favorable performance compared to existing unsupervised and competitive fully supervised methods.   
\end{comment}
In this paper, we break away from traditional unidirectional cross-modal alignment by introducing Structural Semantic Reciprocal Learning (SSRL), a novel unsupervised framework for cross-modal person re-identification. Beyond merely reducing modality discrepancies, SSRL drives a paradigm shift toward structural-semantic reciprocal co-evolution. By synergizing spectrum-invariant fine-grained structural decoupling with closed-loop semantic calibration, our approach uncovers a vital insight for cross-modal retrieval: stable, modality-agnostic spatial topologies can act as an anchor to bootstrap noisy semantic learning, which in turn structurally refines the global feature manifold. Extensive evaluations on SYSU-MM01 and RegDB confirm that this reciprocal closed-loop mechanism not only mitigates pseudo-label noise but also enables unsupervised representations to yield competitive or even superior performance against heavily supervised counterparts.


\section{Supplementary Materials}  

% preamble
\subsection{Extended Implementation Details}
Implemented in PyTorch, our framework adopts an ImageNet-pretrained ResNet-50 backbone with two modality-specific shallow encoders, feeding 2048-D global embeddings into shared blocks. For FSD, a horizontal adaptive pooling layer, instantiated via standard \texttt{AdaptiveAvgPool2d((3,1))}, processes the global feature map to extract $K=3$ part representations. These representations are tightly aligned with pedestrian anatomical structures, namely the upper body, torso, and lower body. Each region then utilizes an independent bottleneck head to generate part embeddings, optimized under three joint objective weights ($\lambda_{p}=0.3$, $\lambda_{pa}=0.1$, and $\lambda_{pc}=0.05$) alongside an adversarial alignment loss ($\lambda_{adv}=0.15$). Optimization operates in two sequential stages: Stage I trains the joint holistic and part-aware embeddings under memory-based supervision, while Stage II initializes from the optimal Stage I checkpoint and activates the CSC semantic prototype and cross-modal consistency constraints ($\lambda_{sp}=\lambda_{sc}=0.1$, $\tau=0.05$). All images are resized to $288 \times 144$. Visible images are augmented with random 10-pixel padding, cropping, horizontal flipping, ImageNet normalization, random erasing ($p=0.5$), and dual-branch appearance channel exchange. Infrared images undergo identical padding, cropping, flipping, normalization, and erasing operators, supplemented by channel-adaptive grayscale transformations ($p=0.5$). Training lasts 50 epochs with a total batch size of 64 (16 instances per identity), utilizing the Adam optimizer ($lr=3.5 \times 10^{-4}$, weight decay $5 \times 10^{-4}$) with a \texttt{StepLR} decay factor of 0.1 every 20 epochs. The modality classifier is optimized via SGD ($lr=0.01$, momentum 0.9, Nesterov). Total iterations per epoch are fixed to 200 for SYSU-MM01 and 100 for RegDB. Unsupervised clustering is performed at each epoch using DBSCAN ($\texttt{min\_samples}=4$) on the extracted feature pool. Modality-specific clustering utilizes $k$-reciprocal Jaccard parameters $k_1=30$ and $k_2=6$. For cross-modal neighbor discovery, the parameter tuple is adjusted to $k_1=40$ and $k_2=32$. The dataset-specific clustering radius $\epsilon$ is specified as 0.6 for SYSU-MM01 across all single- and cross-modal branches. For RegDB, the clustering radius $\epsilon$ is set to 0.6 during the intra-modality clustering stage and adjusted to 0.3 for the cross-modal mutual neighbor discovery phase. The hybrid memory momentum and contrastive temperature are fixed to 0.1 and 0.05, respectively. Evaluation follows the standard all-search and indoor-search protocols for SYSU-MM01 and the visible-to-thermal protocol for RegDB, averaging final metrics over 10 randomized trials. Crucially, during inference, all auxiliary part-aware branches and CSC modules are completely discarded to ensure zero extra computational overhead, executing retrieval matching solely via the $L_2$-normalized 2048-dimensional holistic embedding evaluated by standard Rank-$k$ and mAP.

\subsection{Rationale of Horizontal Adaptive Pooling.}
The proposed Horizontal Adaptive Pooling (HAP) partitions the feature map into three coarse anatomical regions, corresponding to the upper body, torso, and lower body. Although this partition is not explicitly pose-adaptive, it is designed according to the characteristics of surveillance-oriented VI-ReID benchmarks (e.g., SYSU-MM01 and RegDB), where pedestrians are predominantly captured in upright walking or standing poses. Under these conditions, coarse horizontal partitioning provides sufficiently stable anatomical priors while avoiding the additional complexity and potential errors introduced by pose estimation or human parsing under severe cross-modal discrepancies.

Unlike conventional part-based methods that mainly enhance local representation diversity, FSD further projects the three body regions into independent structural subspaces through non-sharing bottleneck heads. This decoupled design prevents localized modality-specific disturbances from propagating across different body regions, enabling each structural primitive to preserve complementary identity cues independently. Consequently, the learned structural primitives serve as reliable spatial anchors that facilitate more robust cross-modal pseudo-label learning and identity association, while remaining resilient to moderate pose variations commonly encountered in surveillance scenarios.

\subsection{Efficiency and Complexity Analysis}

To verify the feasibility of the proposed closed-loop semantic calibration, we compare SSRL with the baseline in terms of model parameters, FLOPs, peak GPU memory, training time, and inference speed. All measurements are conducted on the SYSU-MM01 dataset with input size \(288 \times 144\), batch size 64, and a single NVIDIA GeForce RTX 4090 GPU.

\begin{table}[h]
\centering
\caption{Efficiency comparison between the baseline and SSRL.}
\label{tab:efficiency_comparison}
\small

\resizebox{0.92\linewidth}{!}{%
\begin{tabular}{c|c c c c c}
\hline
\textbf{Method} & \multicolumn{5}{c}{\textbf{Computational Cost}} \\
\cline{2-6}
 & Params (M) & FLOPs (G) & Memory (GB) & Train (ms/iter) & Infer. (ms/img) \\
\hline
Baseline & 72.58 & 662.00 & 10.51 & 214.52 & 3.15 \\
SSRL    & \textbf{70.54} & 662.05 & 10.87 & 217.03 & 3.24 \\
\hline
\end{tabular}%
}
\end{table}

As reported in Table~\ref{tab:efficiency_comparison}, SSRL introduces only negligible computational overhead compared with the baseline despite incorporating both Fine-grained Structural Decoupling (FSD) and Closed-loop Semantic Calibration (CSC). Specifically, the number of parameters is slightly reduced from 72.58M to 70.54M, while the FLOPs remain nearly unchanged (662.00G vs. 662.05G). The peak GPU memory increases marginally from 10.51 GB to 10.87 GB, and the training time rises by only 2.51 ms per iteration (214.52 ms vs. 217.03 ms).

The limited overhead mainly stems from the design of CSC, where the shared semantic prototype bank is reconstructed only once at the end of each epoch based on the updated pseudo-labels, rather than being updated during every mini-batch optimization step. Consequently, the additional computation introduced by prototype aggregation is negligible compared with backbone feature extraction and clustering operations.

Moreover, CSC is only involved during training. During inference, neither DBSCAN clustering nor prototype reconstruction is required, resulting in almost identical deployment efficiency to the baseline. As shown in Table~\ref{tab:efficiency_comparison}, the inference latency increases only slightly from 3.15 ms/img to 3.24 ms/img. These results demonstrate that SSRL achieves substantial retrieval performance gains while maintaining favorable computational efficiency and scalability.

\subsection{Additional Ablation and Qualitative Analysis}
\label{sec:supp_retrieval}
\begin{figure}[]
\centering
\includegraphics[width=0.85\textwidth]{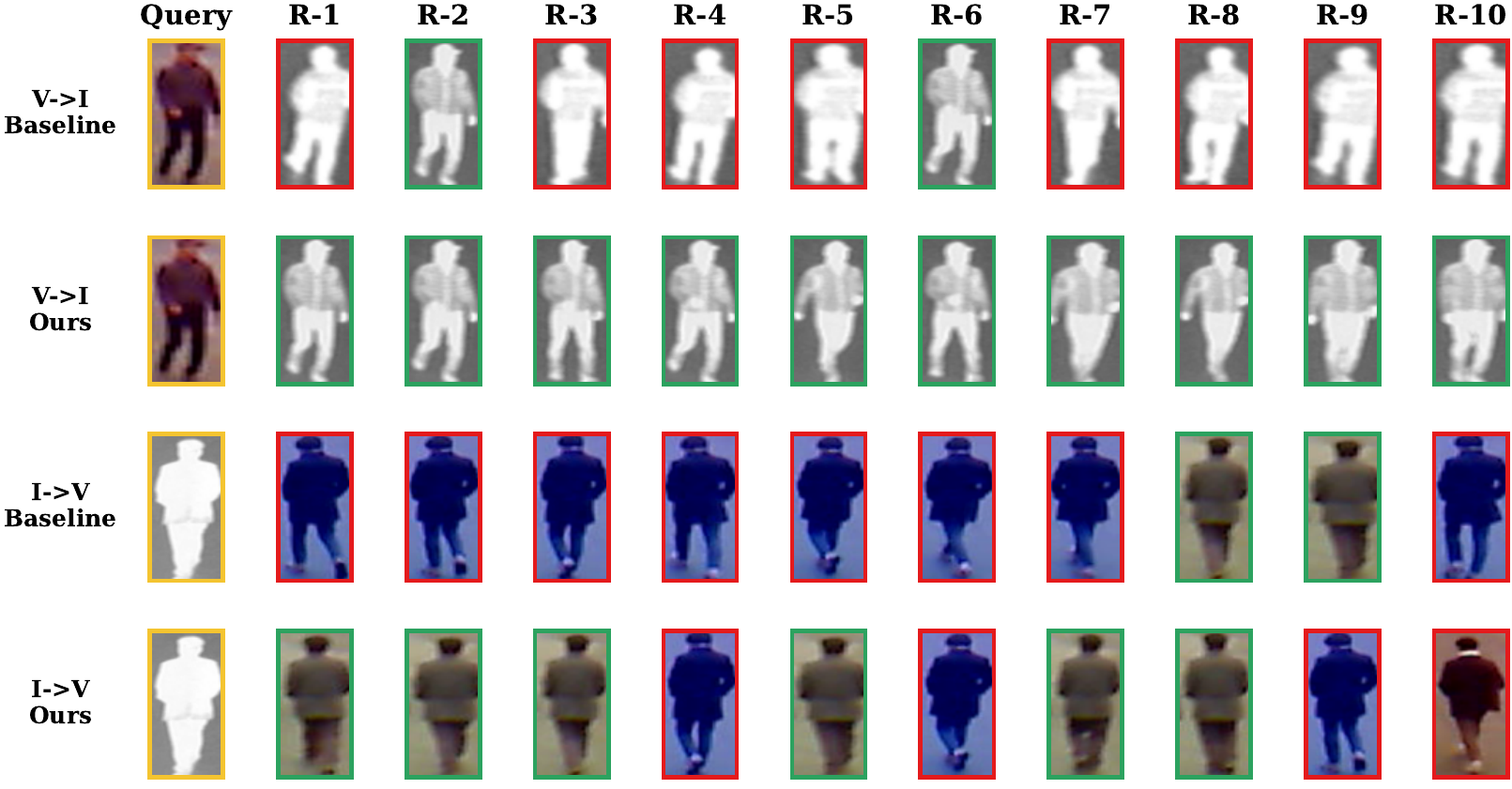}
\caption{Qualitative retrieval comparison on RegDB .}
\label{fig:supp_retrieval_regdb}
\end{figure}

\begin{figure}[]
\centering
\includegraphics[width=0.85\textwidth]{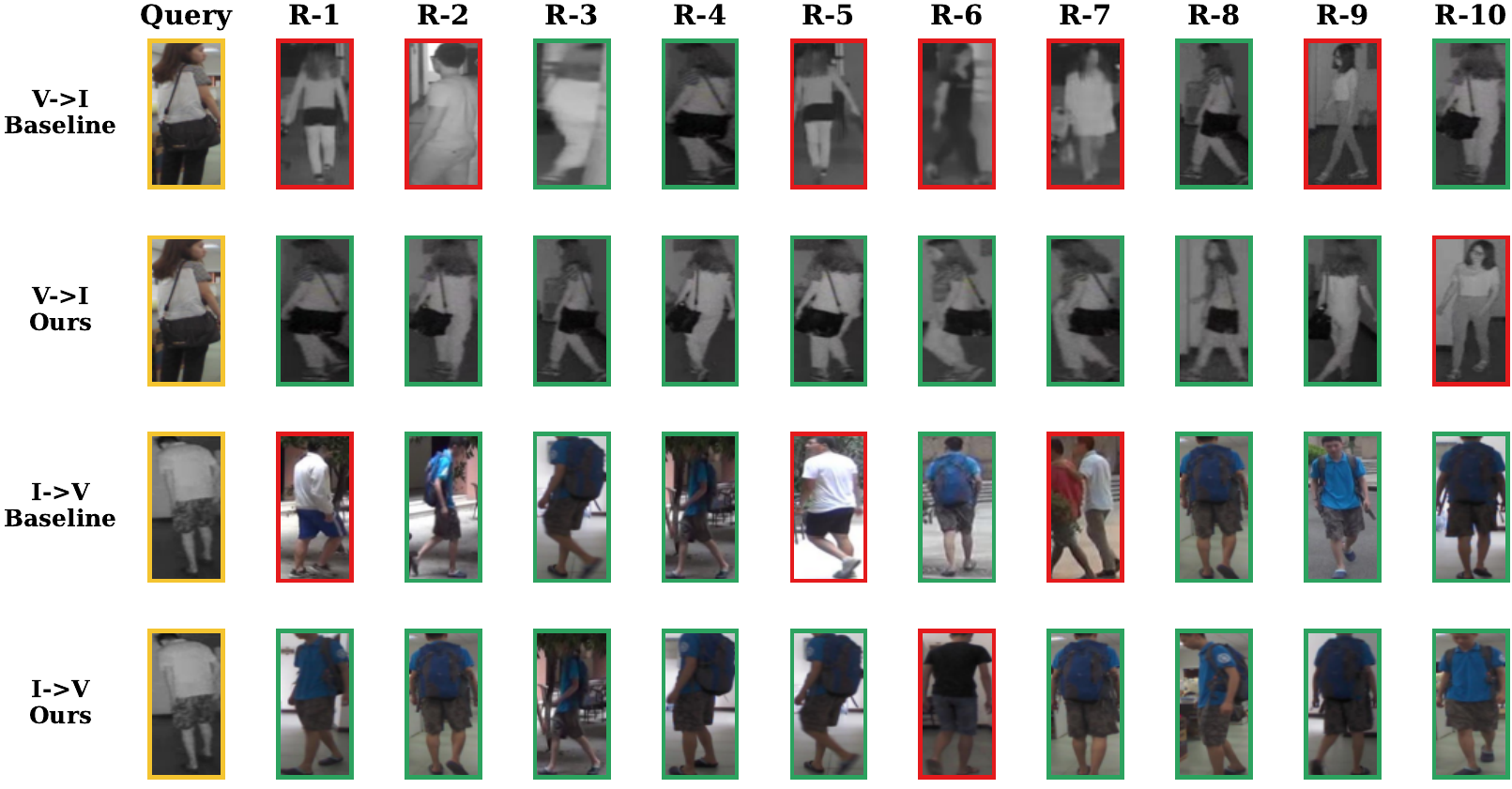}
\caption{Qualitative retrieval comparison on SYSU-MM01 . }
\label{fig:supp_retrieval_sysu}
\end{figure}

\noindent\textbf{Additional Qualitative Retrieval Analysis.}
As illustrated in Fig.~\ref{fig:supp_retrieval_regdb} and Fig.~\ref{fig:supp_retrieval_sysu}, SSRL consistently improves retrieval quality by substantially reducing early-ranking identity confusion.

On RegDB, the baseline suffers from severe cross-modal mismatching. In the $V\!\rightarrow\!I$ setting, false positives appear at Rank-1, Rank-3--5, and Rank-7--10, indicating that the learned representations are highly sensitive to modality discrepancy. In the $I\!\rightarrow\!V$ setting, the baseline produces consecutive errors from Rank-1 to Rank-7 and again at Rank-10, where pedestrians with similar silhouettes and clothing layouts are incorrectly retrieved. By contrast, SSRL achieves a completely correct top-10 ranking in the $V\!\rightarrow\!I$ example and reduces the remaining errors in the $I\!\rightarrow\!V$ example to only Rank-4, Rank-6, Rank-9, and Rank-10.

On SYSU-MM01, the baseline remains vulnerable to viewpoint and background variations. Specifically, false positives occur at Rank-1, Rank-2, Rank-5--7, and Rank-9 in the $V\!\rightarrow\!I$ sequence, where visually similar shoulder bags and upper-body appearances lead to identity confusion. In the $I\!\rightarrow\!V$ sequence, incorrect matches appear at Rank-1, Rank-5, and Rank-7, mainly involving pedestrians with similar backpacks, shorts, and body configurations. In comparison, SSRL retains only one failure at Rank-10 in the $V\!\rightarrow\!I$ example and one failure at Rank-6 in the $I\!\rightarrow\!V$ example, while maintaining correct retrievals at all other positions.
The remaining failures typically occur when different identities share highly similar clothing patterns, accessories, and body structures under substantial viewpoint changes. Nevertheless, most baseline errors are shifted from the top ranks to the tail of the retrieval list, demonstrating the effectiveness of the proposed structural-semantic reciprocal learning strategy in suppressing cross-modal identity confusion.

\noindent\textbf{Sensitivity of k-Reciprocal Neighborhood Size.}
We further investigate the influence of the neighborhood size used in the two-stage clustering pipeline by jointly varying the Stage-I and Stage-II $k_1$ values while keeping all other hyperparameters unchanged.

\begin{table}[h]
\centering
\caption{Sensitivity analysis of the neighborhood size on SYSU-MM01 (\%).}
\label{tab:k_sensitivity}
\small

\resizebox{0.8\linewidth}{!}{%
\begin{tabular}{c|c c c|c c c}
\hline
\multirow{2}{*}{$(k_1^{\mathrm{I}},\,k_1^{\mathrm{II}})$}
& \multicolumn{3}{c|}{\textbf{All Search}}
& \multicolumn{3}{c}{\textbf{Indoor Search}}\\
\cline{2-7}
& Rank-1 & Rank-10 & mAP & Rank-1 & Rank-10 & mAP\\
\hline
(20,30) & 57.06 & 90.89 & 53.39 & 62.95 & 93.95 & 68.65\\
\textbf{(30,40)} & \textbf{59.47} & \textbf{91.84} & \textbf{55.35} & \textbf{63.83} & \textbf{94.57} & \textbf{70.07}\\
(40,50) & 55.73 & 88.83 & 52.27 & 59.21 & 92.62 & 66.27\\
\hline
\end{tabular}%
}
\end{table}

As reported in Table~\ref{tab:k_sensitivity}, the neighborhood configuration has a noticeable impact on pseudo-label quality and the final retrieval performance. The default setting of $(k_1^{\mathrm{I}},k_1^{\mathrm{II}})=(30,40)$ achieves the best overall results, obtaining \textbf{59.47\%} Rank-1 / \textbf{55.35\%} mAP under the All Search protocol and \textbf{63.83\%} Rank-1 / \textbf{70.07\%} mAP under the Indoor Search protocol. This indicates that a moderate neighborhood size provides a better balance between reliable local neighbor estimation and effective cross-modal association.

When smaller neighborhoods are adopted, i.e., $(20,30)$, the retrieval performance decreases due to insufficient neighborhood information for constructing reliable reciprocal relationships, which may introduce noisy cluster assignments. Conversely, excessively large neighborhoods, such as $(40,50)$, also lead to performance degradation, since expanded neighborhoods may incorporate less relevant samples and weaken cluster discriminability. These results demonstrate that SSRL is relatively robust to the neighborhood-size selection, while the adopted configuration provides a favorable trade-off for stable pseudo-label generation.

\noindent\textbf{Sensitivity Analysis of Semantic Temperature.}
We further investigate the influence of the semantic temperature parameter $\tau$ in the closed-loop semantic calibration module on both SYSU-MM01 and RegDB.
As reported in Table~\ref{tab:semantic_temp_sysu_regdb}, retrieval performance is closely related to the sharpness of prototype-based similarity distributions during semantic calibration.

Overall, $\tau=0.05$ provides the best trade-off between retrieval accuracy and optimization stability across different benchmarks and evaluation protocols, achieving \textbf{59.47\%} Rank-1 / \textbf{55.35\%} mAP on SYSU-MM01 All Search and \textbf{89.42\%} mAP under the RegDB Visible$\rightarrow$Infrared setting.
These results indicate that a moderate temperature maintains an appropriate entropy level for prototype assignment while preserving discriminative cross-modal feature alignment.
\label{sec:supp_temperature}
\begin{table}[h]
\centering
\caption{Sensitivity analysis of the semantic temperature \(\tau\) on SYSU-MM01 and RegDB (\%).}
\label{tab:semantic_temp_sysu_regdb}
\small

\resizebox{0.98\linewidth}{!}{%
\begin{tabular}{c|c c|c c|c c|c c}
\hline
\multirow{3}{*}{\textbf{Temperature} \(\tau\)}
& \multicolumn{4}{c|}{\textbf{SYSU-MM01}} 
& \multicolumn{4}{c}{\textbf{RegDB}} \\
\cline{2-9}
& \multicolumn{2}{c|}{All Search} 
& \multicolumn{2}{c|}{Indoor Search} 
& \multicolumn{2}{c|}{Visible$\rightarrow$Infrared} 
& \multicolumn{2}{c}{Infrared$\rightarrow$Visible} \\
\cline{2-9}
& Rank-1 & mAP & Rank-1 & mAP & Rank-1 & mAP & Rank-1 & mAP \\
\hline
0.01 &  48.85&  44.66&  50.10&  57.38& 91.29 & 85.87 &  \textbf{91.22} & 84.88 \\
0.03 &  54.46&  51.56&  58.68&  65.75& 91.85 & 88.06 & 90.24 & 86.61 \\
\textbf{0.05} & \textbf{59.47} & \textbf{55.35} & \textbf{64.83} & \textbf{70.07} & 92.47 & \textbf{89.42} &91.06 & \textbf{88.19} \\
0.07 & 56.12 & 53.31 & 62.20 & 68.69 & \textbf{92.56} & 89.35 & 90.08 & 87.61 \\
0.10 & 58.29 & 54.80 & 61.59 & 68.47 & 90.87 & 87.74 & 88.05 & 85.88 \\
\hline
\end{tabular}%
}
\end{table}
When $\tau$ is excessively small (e.g., $0.01$), the similarity distribution becomes overly concentrated, which may lead to unstable optimization and overconfident prototype matching.
Conversely, larger temperatures (e.g., $0.10$) excessively smooth semantic responses and weaken feature discrimination, resulting in noticeable performance degradation on both datasets.
Although slightly higher Rank-1 scores are observed under certain individual settings (e.g., $\tau=0.07$ for RegDB Visible$\rightarrow$Infrared and $\tau=0.01$ for Infrared$\rightarrow$Visible), their overall mAP performance and cross-setting consistency remain inferior to those achieved with $\tau=0.05$.

Overall, the proposed SSRL framework maintains relatively stable performance within a moderate temperature range, demonstrating the robustness and generalization capability of the proposed closed-loop semantic calibration mechanism under different cross-modal retrieval scenarios.

\subsection{Detailed Training Pipeline of SSRL}
To facilitate reproducibility, Algorithm~\ref{alg:ssrl} outlines the complete training procedure of SSRL. The pipeline consists of two sequential stages, each executed for 50 epochs: Stage I focusing on Feature Structural Decomposition (FSD) optimization, and Stage II dedicated to Cross-modal Semantic Consistency (CSC) learning. This explicit overview clarifies the alternate clustering and mini-batch gradient propagation flow.
\begin{algorithm}[t]
\caption{Two-stage Optimization of SSRL}
\label{alg:ssrl}
\begin{algorithmic}[1]
\REQUIRE Unlabeled visible set $X_v=\{x_v^i\}_{i=1}^{N_v}$, unlabeled infrared set $X_{ir}=\{x_{ir}^j\}_{j=1}^{N_{ir}}$, Stage-I epochs $T_1$, Stage-II epochs $T_2$
\ENSURE Optimized parameters $\theta$

\STATE Initialize network parameters $\theta$, global memory banks $\mathcal{M}_v,\mathcal{M}_{ir},\mathcal{M}_{all}$, and part memory banks $\{\mathcal{M}_v^{pk},\mathcal{M}_{ir}^{pk},\mathcal{M}_{all}^{pk}\}_{k\in\mathcal{K}}$
\STATE Set the region prior $\alpha=[1/6,3/6,2/6]$

\STATE \textbf{Stage I: Fine-grained Structural Decoupling}
\FOR{$t=1$ \TO $T_1$}
    \STATE Extract shared feature maps and structural primitives 
    \STATE Generate modality-specific pseudo-labels for visible and infrared samples by DBSCAN
    \STATE Update modality-specific and shared memory banks
    \FOR{each mini-batch $(B_v,B_{ir})$}
        \STATE Compute holistic and part features by Eq.~\eqref{eq:2}--\eqref{eq:4}
        \STATE Compute $\mathcal{L}_{global}$, $\mathcal{L}_{part}^{uni}$, and $\mathcal{L}_{part}^{all}$ by Eq.~\eqref{eq:5}
        \STATE Compute $\mathcal{L}_{pa}$ and $\mathcal{L}_{pc}$ by Eq.~\eqref{eq:6}--\eqref{eq:7} 
        \STATE Form $\mathcal{L}_{FSD}=\lambda_p\mathcal{L}_{part}+\lambda_{pa}\mathcal{L}_{pa}+\lambda_{pc}\mathcal{L}_{pc}$
        \STATE Compute adversarial alignment loss $\mathcal{L}_{adv}$
        \STATE Update $\theta$ by minimizing
        \[
        \mathcal{L}_{stage1}
        =
        \mathcal{L}_{global}
        +
        \mathcal{L}_{FSD}
        +
        \lambda_{adv}\mathcal{L}_{adv}
        \]
    \ENDFOR
\ENDFOR

\STATE Load the best checkpoint from Stage I

\STATE \textbf{Stage II: Closed-loop Semantic Calibration}
\FOR{$t=1$ \TO $T_2$}
    \STATE Extract holistic features and construct the cross-modal feature set $\mathcal{F}^t=\mathcal{F}_v^t\cup\mathcal{F}_{ir}^t$
    \STATE Compute the modality-invariant k-reciprocal Jaccard distance $D^{J,t}$ by Eq.~\eqref{eq:8}
    \STATE Generate synchronized pseudo-labels $\mathcal{Y}^t=\mathrm{DBSCAN}(D^{J,t})$
    \STATE Update global memory banks and part memory banks using $\mathcal{Y}^t$
    \STATE Construct the shared prototype bank $\mathcal{P}^t$ by Eq.~\eqref{eq:9} 
    \FOR{each mini-batch $(B_v,B_{ir})$}
        \STATE Compute $\mathcal{L}_{global}$ and $\mathcal{L}_{FSD}$ as in Stage I
        \STATE Compute posterior prototype probabilities by Eq.~\eqref{eq:10} 
        \STATE Compute $\mathcal{L}_{sp}$ and $\mathcal{L}_{sc}$ by Eq.~\eqref{eq:11}--\eqref{eq:12}
        \STATE Form $\mathcal{L}_{csc}=\lambda_{sp}\mathcal{L}_{sp}+\lambda_{sc}\mathcal{L}_{sc}$
        \STATE Compute adversarial alignment loss $\mathcal{L}_{adv}$
        \STATE Update $\theta$ by minimizing the overall objective in Eq.~\eqref{eq:13}:
        \[
        \mathcal{L}_{total}^t
        =
        \mathcal{L}_{global}
        +
        \mathcal{L}_{FSD}
        +
        \mathcal{L}_{csc}
        +
        \lambda_{adv}\mathcal{L}_{adv}
        \]
    \ENDFOR
\ENDFOR

\STATE \textbf{return} $\theta$
\end{algorithmic}
\end{algorithm}

\end{document}